\documentclass[10pt,twocolumn,letterpaper]{article}

\usepackage{cvpr}
\usepackage{times}
\usepackage{amsmath}
\usepackage{amssymb}
\usepackage{booktabs}
\usepackage[usenames, dvipsnames]{color}
\usepackage{diagbox}
\usepackage{enumitem}
\usepackage{fancyvrb}
\usepackage{mathtools}
\usepackage{pgfplots}
\usepackage{subfig}
\usepackage{caption}
\usepackage{multicol}
\usepackage{tikz}
\usetikzlibrary{shapes.geometric,arrows,positioning}
\usepackage[]{algorithm}
\usepackage{algpseudocode}
\usepackage{listings}
\usepackage[utf8]{inputenc}

\definecolor{darkgreen}{RGB}{85, 107, 47}
\definecolor{darkred}{RGB}{178, 34, 34}
\definecolor{plotgreen}{RGB}{109,153,162}
\definecolor{plotred}{RGB}{254,135,135}
\definecolor{sourceblue}{rgb}{0,0,0.5}
\definecolor{sourcered}{rgb}{0.6,0,0}
\definecolor{sourcegreen}{rgb}{0,0.5,0}

\usepackage[pagebackref=true,breaklinks=true,letterpaper=true,colorlinks,bookmarks=false]{hyperref}

\cvprfinalcopy 

\def\cvprPaperID{777} 

\newcommand{\defas}{\coloneqq}
\newcommand{\refsec}[1]{Section.~\ref{#1}}
\ifcvprfinal\fi
\numberwithin{table}{section}
\numberwithin{figure}{section}
\DeclareMathOperator{\clamp}{clamp}

\pgfplotsset{compat=1.14}

\newcommand\pythonstyle{\lstset{
language=Python,
basicstyle={\small\ttfamily},
otherkeywords={with},
keywordstyle=\color{sourceblue},
emph={create_training_graph, create_eval_graph},
emphstyle=\color{sourcered},
stringstyle=\color{sourcegreen},
commentstyle=\color{sourcegreen},
frame=tb,
showstringspaces=false
}}

\lstnewenvironment{python}[1][]
{
\pythonstyle
\lstset{#1}
}{}

\begin{document}

\title{Quantization and Training of Neural Networks for Efficient Integer-Arithmetic-Only Inference}

\author{Benoit Jacob\qquad Skirmantas Kligys\qquad Bo Chen\qquad Menglong Zhu\\ Matthew Tang\qquad Andrew Howard\qquad Hartwig Adam\qquad Dmitry Kalenichenko \\
{\tt\small \{benoitjacob,skligys,bochen,menglong,}\\
{\tt\small mttang,howarda,hadam,dkalenichenko\}@google.com}\\
Google Inc.
}

\maketitle

\begin{abstract}
   The rising popularity of intelligent mobile devices and the daunting computational cost of deep learning-based models call for efficient and accurate on-device inference schemes. We propose a quantization scheme that allows inference to be carried out using integer-only arithmetic, which can be implemented more efficiently than floating point inference on commonly available integer-only hardware. We also co-design a training procedure to preserve end-to-end model accuracy post quantization. As a result, the proposed quantization scheme improves the tradeoff between accuracy and on-device latency. The improvements are significant even on MobileNets, a model family known for run-time efficiency, and are demonstrated in ImageNet classification and COCO detection on popular CPUs. 
\end{abstract}

\section{Introduction}

\tikzstyle{nobox}=[text centered, text=blue!60, font=\sffamily\bfseries]
\tikzstyle{act}=[thick, rectangle, rounded corners, minimum width=2cm, minimum height=1cm, text centered, draw=black, fill=yellow!20, font=\sffamily\bfseries\boldmath]
\tikzstyle{add}=[thick, ellipse, minimum width=1cm, minimum height=1cm, text centered, draw=black, fill=yellow!20, font=\sffamily\bfseries\boldmath]
\tikzstyle{conv}=[thick, rectangle, rounded corners, minimum width=2cm, minimum height=1cm, text centered, draw=black, fill=green!20, font=\sffamily\bfseries]
\tikzstyle{params}=[thick, ellipse, minimum width=2cm, minimum height=1cm, text centered, draw=black, fill=blue!20, font=\sffamily\bfseries\boldmath]
\tikzstyle{small_params}=[thick, ellipse, minimum width=1cm, minimum height=1cm, text centered, draw=black, fill=blue!20, font=\sffamily\bfseries\boldmath]
\tikzstyle{quant}=[thick, ellipse, minimum width=2cm, minimum height=1cm, text centered, draw=black, fill=red!20, font=\sffamily\bfseries]
\tikzstyle{ma}=[thick, ellipse, minimum width=2cm, minimum height=1cm, text centered, draw=black, fill=gray!20, font=\sffamily\bfseries\boldmath]
\tikzstyle{arrow} = [thick,->,>=stealth]

\begin{figure*}
    \centering
\subfloat[Integer-arithmetic-only inference]{
\resizebox{0.28\textwidth}{!}{
\begin{tikzpicture}[node distance=2cm,scale=0.8,transform shape]
\node (conv) [conv] {conv};
\node[label={uint8}] (weights) [params, below right of=conv] {weights};
\node (input) [nobox, below left of=conv] {input};
\node (add) [add, above of=conv] {+};
\node[label={uint32}] (biases) [params, below left of=add, xshift=-0.8cm] {biases};
\node (ReLU6) [act, above of=add] {ReLU6};
\node (output) [nobox, right of=ReLU6, xshift=0.8cm] {output};
\draw [arrow] (input) -- (conv) node[midway,left] {uint8};
\draw [arrow] (weights) -- (conv);
\draw [arrow] (conv) -- (add) node[midway,right] {uint32};
\draw [arrow] (biases) -- (add);
\draw [arrow] (add) -- (ReLU6) node[midway,right] {uint8};
\draw [arrow] (ReLU6) -- (output) node[midway,above] {uint8};
\end{tikzpicture}}}%
\subfloat[Training with simulated quantization]{
\resizebox{0.39\textwidth}{!}{
\begin{tikzpicture}[node distance=2cm,scale=0.8,transform shape]
\node (conv) [conv] {conv};
\node (wt_quant) [quant, below right of=conv] {wt quant};
\node (weights) [params, right of=wt_quant, xshift=0.9cm] {weights};
\node (input) [nobox, below left of=conv] {input};
\node (add) [add, above of=conv] {+};
\node (biases) [params, below left of=add, xshift=-1cm] {biases};
\node (ReLU6) [act, above of=add] {ReLU6};
\node (act_quant) [quant, right of=ReLU6, xshift=1.0cm] {act quant};
\node (output) [nobox, right of=act_quant, xshift=0.3cm] {output};
\draw [arrow] (input) -- (conv);
\draw [arrow] (weights) -- (wt_quant);
\draw [arrow] (wt_quant) -- (conv);
\draw [arrow] (conv) -- (add);
\draw [arrow] (biases) -- (add);
\draw [arrow] (add) -- (ReLU6);
\draw [arrow] (ReLU6) -- (act_quant);
\draw [arrow] (act_quant) -- (output);
\end{tikzpicture}}
}%
\subfloat[ImageNet latency-vs-accuracy tradeoff]{
\resizebox{0.3\textwidth}{!}{
\centering
\hspace{-0.3in}
\begin{tikzpicture}
\begin{axis}[
    compat=1.3,
    xlabel={Latency (ms)},
    ylabel={Top $1$ Accuracy},
    xmin=8, xmax=400,
    ymin=40, ymax=72,
    xtick={10, 20, 40, 80, 160, 320},
    ytick={40, 50, 60, 70},
    legend pos=south east,
    ymajorgrids=true,
    xmajorgrids=true,
    xmode=log,
    log ticks with fixed point,
    grid style=dashed,
]
 
\addplot[
    only marks,
    color=plotred,
    mark=*,
    mark options={scale=1.5},
    ]
    coordinates {
(355,70.4)(266,69.3)(198,67.2)(142,64.1)(110,63.6)(82.3,62.1)(62.6,59.9)(42.8,56.2)(39.4,50.4)(30.2,49)(21,46)(14.6,41.3)
    };
\addplot[
    only marks,
    color=plotgreen,
    mark=square*,
    mark options={scale=1.5},
    ]
    coordinates {
(171,69.2)(144,67.9)(90.1,65.8)(59.3,61.8)(60.1,61.3)(49.8,60.2)(31.5,58.4)(21,54.9)(26.1,48.9)(21.3,46.7)(13.6,44.4)(9,40.3)(13.1,50)
};
    \legend{Float, 8-bit}
 
\end{axis}
\end{tikzpicture}
}    
}
    \caption{\textbf{Integer-arithmetic-only quantization.} \small{\textbf{a)} Integer-arithmetic-only inference of a convolution layer. The input and output are represented as $8$-bit integers according to equation \ref{eq:quant-scheme}. The convolution involves 8-bit integer operands and a 32-bit integer accumulator. The bias addition involves only 32-bit integers (section~\ref{sec:fused-layer}). The ReLU6 nonlinearity only involves $8$-bit integer arithmetic. \textbf{b)} Training with simulated quantization of the convolution layer. All variables and computations are carried out using 32-bit floating-point arithmetic. Weight quantization (``wt quant") and activation quantization (``act quant") nodes are injected into the computation graph to simulate the effects of quantization of the variables (section~\ref{sec:training}). The resultant graph approximates the integer-arithmetic-only computation graph in panel a), while being trainable using conventional optimization algorithms for floating point models. \textbf{c)} Our quantization scheme benefits from the fast integer-arithmetic circuits in common CPUs to deliver an improved latency-vs-accuracy tradeoff (section~\ref{sec:experiments}). The figure compares integer quantized MobileNets \cite{MobilenetV1} against floating point baselines on ImageNet \cite{deng2009imagenet} using Qualcomm Snapdragon 835 LITTLE cores.}}
    \label{fig:titlefig}
\end{figure*}
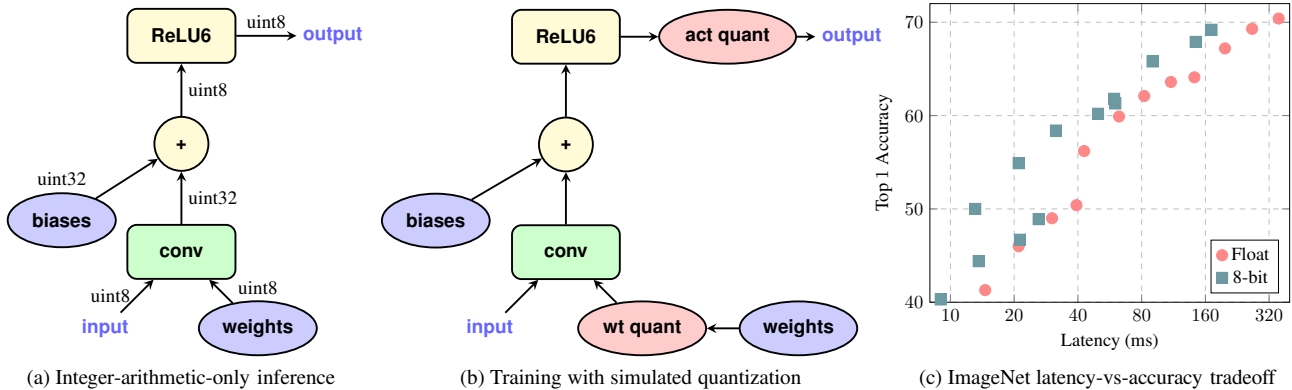

Current state-of-the-art Convolutional Neural Networks (CNNs) are not well suited for use on mobile devices. Since the advent of AlexNet \cite{krizhevsky2012imagenet}, modern CNNs have primarily been appraised according to classification / detection accuracy. Thus network architectures have evolved without regard to model complexity and computational efficiency. On the other hand, successful deployment of CNNs on mobile platforms such as smartphones, AR/VR devices (HoloLens, Daydream), and drones require small model sizes to accommodate limited on-device memory, and low latency to maintain user engagement. This has led to a burgeoning field of research that focuses on reducing the model size and inference time of CNNs with minimal accuracy losses. 

Approaches in this field roughly fall into two categories. The first category, exemplified by MobileNet \cite{MobilenetV1}, SqueezeNet \cite{iandola2016squeezenet}, ShuffleNet \cite{ShuffleNet2017}, and DenseNet \cite{Huang_2017_CVPR}, designs novel network architectures that exploit computation / memory efficient operations. The second category quantizes the weights and / or activations of a CNN from 32 bit floating point into lower bit-depth representations. This methodology, embraced by approaches such as Ternary weight networks (TWN \cite{li2016ternary}), Binary Neural Networks (BNN \cite{hubara2016binarized}), XNOR-net \cite{rastegari2016xnor}, and more \cite{han2015deep,leng2017extremely,mellempudi2017ternary,zhou2017incremental,zhou2016dorefa,zhu2016trained}, is the focus of our investigation. Despite their abundance, current quantization approaches are lacking in two respects when it comes to trading off latency with accuracy.

First, prior approaches have not been evaluated on a reasonable baseline architecture. The most common baseline architectures, AlexNet \cite{krizhevsky2012imagenet}, VGG \cite{simonyan2014very} and GoogleNet \cite{szegedy2015going}, are all over-parameterized by design in order to extract marginal accuracy improvements. Therefore, it is easy to obtain sizable compression of these architectures, reducing quantization experiments on these architectures to proof-of-concepts at best. Instead, a more meaningful challenge would be to quantize model architectures that are already efficient at trading off latency with accuracy, e.g. MobileNets.

Second, many quantization approaches do not deliver verifiable efficiency improvements on real hardware. Approaches that quantize only the weights (\cite{chen2015compressing,gong2014compressing,han2015deep,zhou2017incremental}) are primarily concerned with on-device storage and less with computational efficiency. Notable exceptions are binary, ternary and bit-shift networks~\cite{hubara2016binarized,li2016ternary,rastegari2016xnor}. These latter approaches employ weights that are either 0 or powers of 2, which allow multiplication to be implemented by bit shifts. However, while bit-shifts can be efficient in custom hardware, they provide little benefit on existing hardware with multiply-add instructions that, when properly used (i.e. pipelined), are not more expensive than additions alone. Moreover, multiplications are only expensive if the operands are wide, and the need to avoid multiplications diminishes with bit depth once both weights and activations are quantized. Notably, these approaches rarely provide on-device measurements to verify the promised timing improvements. More runtime-friendly approaches quantize both the weights and the activations into 1 bit representations \cite{hubara2016binarized,rastegari2016xnor,zhou2016dorefa}. With these approaches, both multiplications and additions can be implemented by efficient bit-shift and bit-count operations, which are showcased in custom GPU kernels (BNN \cite{hubara2016binarized}). However, 1 bit quantization often leads to substantial performance degradation, and may be overly stringent on model representation.

In this paper we address the above issues by improving the latency-vs-accuracy tradeoffs of MobileNets on common mobile hardware. Our specific contributions are: 
\begin{itemize}[leftmargin=*]
    \setlength\itemsep{0em}
\item We provide a \emph{quantization scheme} (section \ref{sec:quant-scheme}) that quantizesh both weights and activations as 8-bit integers, and just a few parameters (bias vectors) as 32-bit integers. 
\item We provide a \emph{quantized inference framework} that is efficiently implementable on integer-arithmetic-only hardware such as the Qualcomm Hexagon (sections \ref{sec:integer-only-inference}, \ref{sec:matrix-multiplication}), and we describe an efficient, accurate implementation on ARM NEON (Appendix~\ref{sec:arm-neon}). 
\item We provide a \emph{quantized training framework} (section \ref{sec:training}) co-designed with our quantized inference to minimize the loss of accuracy from quantization on real models. 
\item We apply our frameworks to efficient classification and detection systems based on MobileNets and provide benchmark results on popular ARM CPUs (section \ref{sec:experiments}) that show \emph{significant improvements} in the latency-vs-accuracy tradeoffs for state-of-the-art MobileNet architectures, demonstrated in ImageNet classification \cite{deng2009imagenet}, COCO object detection \cite{lin2014microsoft}, and other tasks.
\end{itemize}

Our work draws inspiration from \cite{gupta2015deep}, which leverages low-precision fixed-point arithmetic to accelerate the training speed of CNNs, and from \cite{vanhoucke2011improving}, which uses $8$-bit fixed-point arithmetic to speed up inference on x86 CPUs. Our quantization scheme focuses instead on improving the inference speed vs accuracy tradeoff on mobile CPUs. 

\section{Quantized Inference}\label{sec:quantized-inference}

\subsection{Quantization scheme}\label{sec:quant-scheme}
In this section, we describe our general \emph{quantization scheme}\footnote{The quantization scheme described here is the one adopted in TensorFlow Lite \cite{tflite} and we will refer to specific parts of its code to illustrate aspects discussed below.}\footnote{We had earlier described this quantization scheme in the \href{https://github.com/google/gemmlowp/blob/fcf32e7a0a4d2af46e63eccf0c8fa4d83d0311c5/doc/quantization.md}{documentation} of gemmlowp \cite{gemmlowp}. That page may still be useful as an alternate treatment of some of the topics developed in this section, and for its self-contained example code.}, that is, the correspondence between the bit-representation of values (denoted $q$ below, for ``quantized value") and their interpretation as mathematical real numbers (denoted $r$ below, for ``real value"). Our quantization scheme is implemented using integer-only arithmetic during inference and floating-point arithmetic during training, with both implementations maintaining a high degree of correspondence with each other. We achieve this by first providing a mathematically rigorous definition of our quantization scheme, and separately adopting this scheme for both integer-arithmetic inference and floating-point training.

A basic requirement of our quantization scheme is that it permits efficient implementation of all arithmetic using only integer arithmetic operations on the quantized values (we eschew implementations requiring lookup tables because these tend to perform poorly compared to pure arithmetic on SIMD hardware). This is equivalent to requiring that the quantization scheme be an \emph{affine mapping} of integers $q$ to real numbers $r$, i.e. of the form
\begin{equation}\label{eq:quant-scheme}
r = S(q - Z)
\end{equation}
for some constants $S$ and $Z$. Equation (\ref{eq:quant-scheme}) is our \emph{quantization scheme} and the constants $S$ and $Z$ are our \emph{quantization parameters}. Our quantization scheme uses a single set of quantization parameters for all values within each activations array and within each weights array; separate arrays use separate quantization parameters.

For \emph{8-bit quantization}, $q$ is quantized as an 8-bit integer (for $B$-bit quantization, $q$ is quantized as an $B$-bit integer). Some arrays, typically bias vectors, are quantized as 32-bit integers, see section \ref{sec:fused-layer}.

The constant $S$ (for ``scale") is an arbitrary positive real number. It is typically represented in software as a floating-point quantity, like the real values $r$. Section \ref{sec:integer-only-inference} describes methods for avoiding the representation of such floating-point quantities in the inference workload.

The constant $Z$ (for ``zero-point") is of the same type as quantized values $q$, and is in fact the quantized value $q$ corresponding to the real value 0. This allows us to automatically meet the requirement that the real value $r=0$ be exactly representable by a quantized value. The motivation for this requirement is that efficient implementation of neural network operators often requires zero-padding of arrays around boundaries.

Our discussion so far is summarized in the following \emph{quantized buffer} data structure\footnote{The actual data structures in the TensorFlow Lite \cite{tflite} Converter are \texttt{QuantizationParams} and \texttt{Array} in \href{https://github.com/tensorflow/tensorflow/blob/4952f981be07b8bf508f8226f83c10cdafa3f0c4/tensorflow/contrib/lite/toco/model.h}{this header file}. As we discuss in the next subsection, this data structure, which still contains a floating-point quantity, does not appear in the actual quantized on-device inference code.}, with one instance of such a buffer existing for each activations array and weights array in a neural network. We use C++ syntax because it allows the unambiguous conveyance of types.
{
\begin{Verbatim}[fontsize=\footnotesize]
template<typename QType>  // e.g. QType=uint8
struct QuantizedBuffer {
  vector<QType> q;        // the quantized values
  float S;                // the scale
  QType Z;                // the zero-point
};
\end{Verbatim}
}

\subsection{Integer-arithmetic-only matrix multiplication}\label{sec:integer-only-inference}

We now turn to the question of how to perform inference using only integer arithmetic, i.e. how to use Equation (\ref{eq:quant-scheme}) to translate real-numbers computation into quantized-values computation, and how the latter can be designed to involve only integer arithmetic even though the scale values $S$ are not integers.

Consider the multiplication of two square $N\times N$ matrices of real numbers, $r_1$ and $r_2$, with their product represented by $r_3 = r_1r_2$. We denote the entries of each of these matrices $r_\alpha$ ($\alpha=1$, $2$ or $3$) as $r_\alpha^{(i,j)}$ for $1\leqslant i,j\leqslant N$, and the quantization parameters with which they are quantized as $(S_\alpha, Z_\alpha)$. We denote the quantized entries by $q_\alpha^{(i,j)}$. Equation (\ref{eq:quant-scheme}) then becomes:
\begin{equation}
\label{eq:scalar-operands}
r_\alpha^{(i,j)} = S_\alpha(q_\alpha^{(i,j)} - Z_\alpha).
\end{equation}
From the definition of matrix multiplication, we have
\begin{equation}
\label{eq:quant-product-1}
S_3(q_3^{(i,k)}-Z_3) = \sum_{j=1}^N S_1(q_1^{(i,j)} - Z_1) S_2(q_2^{(j,k)} - Z_2),
\end{equation}
which can be rewritten as
\begin{equation}
\label{eq:quant-product-2}
q_3^{(i, k)} = Z_3 + M \sum_{j=1}^N (q_1^{(i, j)} - Z_1) (q_2^{(j, k)} - Z_2),
\end{equation}
where the \emph{multiplier} $M$ is defined as
\begin{equation}
\label{eq:multiplier}
M \defas \frac{S_1 S_2}{S_3}.
\end{equation}
In Equation (\ref{eq:quant-product-2}), the only non-integer is the multiplier $M$. As a constant depending only on the quantization scales $S_1, S_2, S_3$, it can be computed offline. We empirically find it to always be in the interval $(0, 1)$, and can therefore express it in the normalized form
\begin{equation}
M = 2^{-n} M_0
\end{equation}
where $M_0$ is in the interval $[0.5, 1)$ and $n$ is a non-negative integer. The normalized multiplier $M_0$ now lends itself well to being expressed as a fixed-point multiplier (e.g. 
int16 or int32 depending on hardware capability). For example, if int32 is used, the integer representing $M_0$ is the int32 value nearest to $2^{31}M_0$. Since $M_0\geqslant 0.5$, 
this value is always at least $2^{30}$ and will therefore always have at least 30 bits of relative accuracy. Multiplication by $M_0$ can thus be implemented as a fixed-point 
multiplication\footnote{The computation discussed in this section is implemented in TensorFlow Lite 
\cite{tflite}\href{https://github.com/tensorflow/tensorflow/blob/4952f981be07b8bf508f8226f83c10cdafa3f0c4/tensorflow/contrib/lite/kernels/internal/reference/reference_ops.h\#L493-L534}{
reference code} for a fully-connected layer.}. Meanwhile, multiplication by $2^{-n}$ can be implemented with an efficient bit-shift, albeit one that needs to have correct 
round-to-nearest behavior, an issue that we return to in Appendix~\ref{sec:arm-neon}.

\subsection{Efficient handling of zero-points}
\label{sec:matrix-multiplication}

In order to efficiently implement the evaluation of Equation (\ref{eq:quant-product-2}) without having to perform $2N^3$ subtractions and without having to expand the operands of the multiplication into 16-bit integers, we first notice that by distributing the multiplication in Equation (\ref{eq:quant-product-2}), we can rewrite it as
\begin{equation}
\label{eq:quant-product-matrix-expanded}
\begin{split}
q_3^{(i, k)} = Z_3 + M \left( \vphantom{\sum_{j=1}^N} NZ_1 Z_2 - Z_1 a_2^{(k)} \right. \\
\left. - Z_2 \bar a_1^{(i)} + \sum_{j=1}^N q_1^{(i, j)}q_2^{(j, k)}\right)
\end{split}
\end{equation}
where
\begin{equation}
\label{eq:sums-slices}
a_2^{(k)} \defas \sum_{j=1}^N q_2^{(j, k)},\;\;
\bar a_1^{(i)} \defas \sum_{j=1}^N q_1^{(i, j)}.
\end{equation}
Each $a_2^{(k)}$ or $\bar a_1^{(i)}$ takes only $N$ additions to compute, so they collectively take only $2N^2$ additions. The rest of the cost of the evaluation of (\ref{eq:quant-product-matrix-expanded}) is almost entirely concentrated in the core integer matrix multiplication accumulation
\begin{equation}
\label{eq:core-accumulation}
\sum_{j=1}^N q_1^{(i, j)}q_2^{(j, k)}
\end{equation}
which takes $2N^3$ arithmetic operations; indeed, everything else involved in (\ref{eq:quant-product-matrix-expanded}) is $O(N^2)$ with a small constant in the $O$. Thus, the expansion into the form (\ref{eq:quant-product-matrix-expanded}) and the factored-out computation of $a_2^{(k)}$ and $\bar a_1^{(i)}$ enable low-overhead handling of arbitrary zero-points for anything but the smallest values of $N$, reducing the problem to the same core integer matrix multiplication accumulation (\ref{eq:core-accumulation}) as we would have to compute in any other zero-points-free quantization scheme.

\subsection{Implementation of a typical fused layer}\label{sec:fused-layer}

We continue the discussion of section \ref{sec:matrix-multiplication}, but now explicitly define the data types of all quantities involved, and modify the quantized matrix multiplication (\ref{eq:quant-product-matrix-expanded}) to merge the bias-addition and activation function evaluation directly into it. This fusing of whole layers into a single operation is not only an optimization. As we must reproduce in inference code the same arithmetic that is used in training, the granularity of fused operators in inference code (taking an 8-bit quantized input and producing an 8-bit quantized output) must match the placement of ``fake quantization" operators in the training graph (section \ref{sec:training}).

For our implementation on ARM and x86 CPU architectures, we use the gemmlowp library \cite{gemmlowp}, whose \texttt{GemmWithOutputPipeline} entry point provides supports the fused operations that we now describe\footnote{The discussion in this section is implemented in TensorFlow Lite \cite{tflite} for e.g. a Convolutional operator (\href{https://github.com/tensorflow/tensorflow/blob/4952f981be07b8bf508f8226f83c10cdafa3f0c4/tensorflow/contrib/lite/kernels/internal/reference/reference_ops.h\#L248-L314}{reference code} is self-contained, \href{https://github.com/tensorflow/tensorflow/blob/4952f981be07b8bf508f8226f83c10cdafa3f0c4/tensorflow/contrib/lite/kernels/internal/optimized/optimized_ops.h\#L837-L906}{optimized code} calls into gemmlowp \cite{gemmlowp}).}.

We take the $q_1$ matrix to be the weights, and the $q_2$ matrix to be the activations. Both the weights and activations are of type uint8 (we could have equivalently chosen int8, with suitably modified zero-points). Accumulating products of uint8 values requires a 32-bit accumulator, and we choose a signed type for the accumulator for a reason that will soon become clear. The sum in (\ref{eq:core-accumulation}) is thus of the form:
\begin{equation}
\label{eq:accum-types}
\texttt{int32 += uint8 * uint8}.
\end{equation}
In order to have the quantized bias-addition be the addition of an int32 bias into this int32 accumulator, the bias-vector is quantized such that: it uses int32 as its quantized data type; it uses 0 as its quantization zero-point $Z_{\mathrm{bias}}$; and its quantization scale $S_{\mathrm{bias}}$ is the same as that of the accumulators, which is the product of the scales of the weights and of the input activations. In the notation of section \ref{sec:matrix-multiplication},
\begin{equation}
S_{\mathrm{bias}} = S_1 S_2,\;\;Z_{\mathrm{bias}}=0.
\end{equation}
Although the bias-vectors are quantized as 32-bit values, they account for only a tiny fraction of the parameters in a neural network. Furthermore, the use of higher precision for bias vectors meets a real need: as each bias-vector entry is added to many output activations, any quantization error in the bias-vector tends to act as an overall bias (i.e. an error term with nonzero mean), which must be avoided in order to preserve good end-to-end neural network accuracy\footnote{The quantization of bias-vectors discussed here is implemented \href{https://github.com/tensorflow/tensorflow/blob/4952f981be07b8bf508f8226f83c10cdafa3f0c4/tensorflow/contrib/lite/toco/graph_transformations/quantize.cc\#L171-L197}{here} in the TensorFlow Lite \cite{tflite} Converter.}.

With the final value of the int32 accumulator, there remain three things left to do: \textit{scale down} to the final scale used by the 8-bit output activations, \textit{cast down} to uint8 and \textit{apply the activation function} to yield the final 8-bit output activation.

The down-scaling corresponds to multiplication by the multiplier $M$ in equation (\ref{eq:quant-product-matrix-expanded}). As explained in section \ref{sec:integer-only-inference}, it is implemented as a fixed-point multiplication by a normalized multiplier $M_0$ and a rounding bit-shift. Afterwards, we perform a saturating cast to uint8, saturating to the range $[0, 255]$.

We focus on activation functions that are mere clamps, e.g. ReLU, ReLU6. Mathematical functions are discussed in appendix \ref{sec:math-functions} and we do not currently fuse them into such layers. Thus, the only thing that our fused activation functions need to do is to further clamp the uint8 value to some sub-interval of $[0, 255]$ before storing the final uint8 output activation. In practice, the quantized training process (section \ref{sec:training}) tends to learn to make use of the whole output uint8 $[0, 255]$ interval so that the activation function no longer does anything, its effect being subsumed in the clamping to $[0, 255]$ implied in the saturating cast to uint8.

\section{Training with simulated quantization}\label{sec:training}

A common approach to training quantized networks is to train in floating point and then quantize the resulting weights (sometimes with additional post-quantization training for fine-tuning). We found that this approach works sufficiently well for large models with considerable representational capacity, but leads to significant accuracy drops for small models. Common failure modes for simple post-training quantization include: 1) large differences (more than $100\times$) in \textit{ranges of weights} for different output channels (section~\ref{sec:quantized-inference} mandates that all channels of the same layer be quantized to the same resolution, which causes weights in channels with smaller ranges to have much higher relative error) and 2) \textit{outlier weight values} that make all remaining weights less precise after quantization.

We propose an approach that simulates quantization effects in the forward pass of training. Backpropagation still happens as usual, and all weights and biases  are \textit{stored in floating point} so that they can be easily nudged by small amounts. The forward propagation pass however \textit{simulates quantized inference} as it will happen in the inference engine, by implementing in floating-point arithmetic the rounding behavior of the quantization scheme that we introduced in section \ref{sec:quantized-inference}:

\begin{itemize}[leftmargin=*]
    \setlength\itemsep{0em}
  \item Weights are quantized before they are convolved with the input. If batch normalization (see \cite{batch_norm}) is used for the layer, the batch normalization parameters are ``folded into" the weights before quantization, see section \ref{sec:bn_fold}.
  \item Activations are quantized at points where they would be during inference, e.g. after the activation function is applied to a convolutional or fully connected layer's output, or after a bypass connection adds or concatenates the outputs of several layers together such as in ResNets.
\end{itemize}

For each layer, quantization is parameterized by the number of quantization levels and clamping range, and is performed by applying point-wise the quantization function $q$ defined as follows:
\begin{align}
\clamp(r;a,b) &\defas \min\left(\max(x, a), b\right) \nonumber \\
s(a,b,n) &\defas \frac{b - a}{n - 1} \nonumber \\
q(r;a,b,n) &\defas \left\lfloor \frac{\clamp(r;a,b) - a}{s(a,b,n)} \right\rceil s(a,b,n) + a,\label{eq:fakequant}
\end{align}
where $r$ is a real-valued number to be quantized, $[a;b]$ is the quantization range, $n$ is the number of quantization levels, and $\lfloor\cdot\rceil$ denotes rounding to the nearest integer. $n$ is fixed for all layers in our experiments, e.g. $n = 2^8=256$ for 8 bit quantization.

\subsection{Learning quantization ranges}\label{sec:quant-range}
Quantization ranges are treated differently for weight quantization vs. activation quantization:
\begin{itemize}[leftmargin=*]
    \setlength\itemsep{0em}
  \item For weights, the basic idea is simply to set $a \defas \min w$, $b \defas \max w$. We apply a minor tweak to this so that the weights, once quantized as int8 values, only range in $[-127,127]$ and never take the value $-128$, as this enables a substantial optimization opportunity (for more details, see Appendix~\ref{sec:arm-neon}).
  \item For activations, ranges depend on the inputs to the network. To estimate the ranges, we collect $[a; b]$ ranges seen on activations during training and then aggregate them via exponential moving averages (EMA) with the smoothing parameter being close to 1 so that observed ranges are smoothed across thousands of training steps. Given significant delay in the EMA updating activation ranges when the ranges shift rapidly, we found it useful to completely disable activation quantization at the start of training (say, for 50 thousand to 2 million steps). This allows the network to enter a more stable state where activation quantization ranges do not exclude a significant fraction of values.
\end{itemize}

In both cases, the boundaries $[a; b]$ are nudged so that value $0.0$ is exactly representable as an integer $z(a,b,n)$ after quantization. As a result, the learned quantization parameters map to the scale $S$ and zero-point $Z$ in equation~\ref{eq:quant-scheme}:
\begin{equation}
S= s(a,b,n), \,\,\, Z=z(a,b,n) \label{eq:quant-param-mapping}
\end{equation}

Below we depict simulated quantization assuming that the computations of a neural network are captured as a TensorFlow graph \cite{abadi2015tensorflow}. A typical workflow is described in Algorithm \ref{algo:training}.
\begin{algorithm}
\caption{Quantized graph training and inference}
\begin{algorithmic}[1]
  \State Create a training graph of the floating-point model.
  \State Insert \textit{fake quantization} TensorFlow operations in locations where tensors will be downcasted to fewer bits during inference according to equation \ref{eq:fakequant}.
  \State Train in simulated quantized mode until convergence.
  \State Create and optimize the inference graph for running in a low bit inference engine.
  \State Run inference using the quantized inference graph.
\end{algorithmic}
\label{algo:training}
\end{algorithm}
Optimization of the inference graph by fusing and removing operations is outside the scope of this paper. Source code for graph modifications (inserting \textit{fake quantization} operations, creating and optimizing the inference graph) and a low bit inference engine has been open-sourced with TensorFlow contributions in \cite{tf_quant}. 

Figure \ref{fig:titlefig}a and b illustrate TensorFlow graphs before and after quantization for a simple convolutional layer. Illustrations of the more complex convolution with a bypass connection in figure \ref{fig:bypass_original} can be found in figure \ref{fig:bypass_quantized}.

Note that the biases are not quantized because they are represented as 32-bit integers in the inference process, with a much higher range and precision compared to the 8 bit weights and activations. Furthermore, quantization parameters used for biases are inferred from the quantization parameters of the weights and activations. See section \ref{sec:fused-layer}.

Typical TensorFlow code illustrating use of \cite{tf_quant} follows:


\begin{python}
from tf.contrib.quantize \
    import quantize_graph as qg

g = tf.Graph()
with g.as_default():
  output = ...
  total_loss = ...
  optimizer = ...
  train_tensor = ...
if is_training:
  quantized_graph = \
      qg.create_training_graph(g)
else:
  quantized_graph = \
      qg.create_eval_graph(g)
# Train or evaluate quantized_graph.
\end{python}

\subsection{Batch normalization folding}\label{sec:bn_fold}

For models that use batch normalization (see \cite{batch_norm}), there is additional complexity: the training graph contains batch normalization as a separate block of operations, whereas the inference graph has batch normalization parameters ``folded'' into the convolutional or fully connected layer's weights and biases, for efficiency. To accurately simulate quantization effects, we need to simulate this folding, and quantize weights after they have been scaled by the batch normalization parameters. We do so with the following:
\begin{equation}
w_{\mathrm{fold}} \defas \frac{\gamma w}{\sqrt{EMA(\sigma^2_B) + \varepsilon}}.
\end{equation}
Here $\gamma$ is the batch normalization's scale parameter, $EMA(\sigma^2_B)$ is the moving average estimate of the variance of convolution results across the batch, and $\varepsilon$ is just a small constant for numerical stability.

After folding, the batch-normalized convolutional layer reduces to the simple convolutional layer depicted in figure~\ref{fig:titlefig}a with the folded weights $w_{\mathrm{fold}}$ and the corresponding folded biases. Therefore the same recipe in figure~\ref{fig:titlefig}b applies. See the appendix for the training graph (figure~\ref{fig:bn_train}) for a batch-normalized convolutional layer, the corresponding inference graph (figure~\ref{fig:bn_inference}), the training graph after batch-norm folding (figure~\ref{fig:bn_fold_train}) and the training graph after both folding and quantization (figure~\ref{fig:bn_fold_quant_train}).

\section{Experiments}\label{sec:experiments}
We conducted two set of experiments, one showcasing the effectiveness of quantized training (\refsec{sec:quant-large-nets}), and the other illustrating the improved latency-vs-accuracy tradeoff of quantized models on common hardware (\refsec{sec:quant-mobilenets}). The most performance-critical part of the inference workload on the neural networks being benchmarked is matrix multiplication (GEMM). The 8-bit and 32-bit floating-point GEMM inference code uses the gemmlowp library \cite{gemmlowp} for 8-bit quantized inference, and the Eigen library \cite{eigen} for 32-bit floating-point inference.

\subsection{Quantized training of Large Networks} \label{sec:quant-large-nets}
We apply quantized training to ResNets \cite{he2016deep} and InceptionV3 \cite{szegedy2016rethinking} on the ImageNet dataset. These popular networks are too computationally intensive to be deployed on mobile devices, but are included for comparison purposes. Training protocols are discussed in Appendix~\ref{sec:resnet-protocol} and~\ref{sec:inception-protocol}.

\subsubsection{ResNets}
We compare floating-point vs integer-quantized ResNets for various depths in table \ref{tab:resnet_cmp1}. Accuracies of integer-only quantized networks are within $2\%$ of their floating-point counterparts.

\begin{table}
\centering
\resizebox{\columnwidth}{!}{
\begin{tabular}{cccc}
\toprule
ResNet depth & 50 & 100 & 150 \\
\midrule
Floating-point accuracy & 76.4\% & 78.0\% & 78.8\% \\
Integer-quantized accuracy & 74.9\% & 76.6\% & 76.7\% \\
\bottomrule
\end{tabular}
}
\caption{ResNet on ImageNet: Floating-point vs quantized network accuracy for various network depths.}

\vspace*{9pt}
\label{tab:resnet_cmp1}
\end{table}

We also list ResNet50 accuracies under different quantization schemes in table \ref{tab:resnet_cmp2}. As expected, integer-only quantization outperforms FGQ \cite{mellempudi2017ternary}, which uses 2 bits for weight quantization. INQ \cite{zhou2017incremental} (5-bit weight floating-point activation) achieves a similar accuracy as ours, but we provide additional run-time improvements (see section \ref{sec:quant-mobilenets}).

\begin{table}
\centering
\resizebox{\columnwidth}{!}{
\begin{tabular}{cccccccc}
\toprule
Scheme & BWN & TWN & INQ & FGQ & Ours  \\
\midrule
Weight bits & 1		& 2		& 5		& 2       &8\\
\midrule
Activation bits & float32		& float32		& float32		&       8 & 8\\
\midrule
Accuracy & 68.7\% & 72.5\% & 74.8\% & 70.8\% & 74.9\%  \\
\bottomrule
\end{tabular}
}
\caption{ResNet on ImageNet: Accuracy under various quantization schemes, including binary weight networks (BWN \cite{leng2017extremely,hubara2016quantized}), ternary weight networks (TWN \cite{leng2017extremely,li2016ternary}), incremental network quantization (INQ \cite{zhou2017incremental}) and fine-grained quantization (FGQ \cite{mellempudi2017ternary})}
\label{tab:resnet_cmp2}
\end{table}

\subsubsection{Inception v3 on ImageNet}
We compare the Inception v3 model quantized into 8 and 7 bits, respectively. 7-bit quantization is obtained by setting the number of quantization levels in equation~\ref{eq:fakequant} to $n=2^7$. We additionally probe the sensitivity of activation quantization by comparing networks with two activation nonlinearities, ReLU6 and ReLU. The training protocol is in Appendix~\ref{sec:inception-protocol}.

Table \ref{tab:inception_v3_imagenet} shows that 7-bit quantized training produces model accuracies close to that of 8-bit quantized training, and quantized models with ReLU6 have less accuracy degradation. The latter can be explained by noticing that ReLU6 introduces the interval $[0, 6]$ as a natural range for activations, while ReLU allows activations to take values from a possibly larger interval, with different ranges in different channels. Values in a fixed range are easier to quantize with high precision.

\begin{table}
\centering
\begin{tabular}{cccccc}
\toprule
    Act. & type & \multicolumn{2}{c}{accuracy} & \multicolumn{2}{c}{recall 5} \\
    \cmidrule{3-6}
           &        & mean   & std. dev.           & mean    & std.dev. \\
\midrule
    ReLU6 & floats  & 78.4\% & 0.1\%               & 94.1\%  & 0.1\%    \\
          & 8 bits  & 75.4\% & 0.1\%               & 92.5\%  & 0.1\%    \\
          & 7 bits  & 75.0\% & 0.3\%               & 92.4\%  & 0.2\%    \\
\midrule
    ReLU  & floats  & 78.3\% & 0.1\%               & 94.2\%  & 0.1\%    \\
          & 8 bits  & 74.2\% & 0.2\%               & 92.2\%  & 0.1\%    \\
          & 7 bits  & 73.7\% & 0.3\%               & 92.0\%  & 0.1\%    \\
\bottomrule
\end{tabular}
\vspace*{8pt plus 3pt minus 2pt}
\caption{Inception v3 on ImageNet: Accuracy and recall 5 comparison of floating point and quantized models.}
\label{tab:inception_v3_imagenet}
\end{table}

\subsection{Quantization of MobileNets} \label{sec:quant-mobilenets}
MobileNets are a family of architectures that achieve a state-of-the-art tradeoff between on-device latency and ImageNet classification accuracy. In this section we demonstrate how integer-only quantization can further improve the tradeoff on common hardware.

\subsubsection{ImageNet}
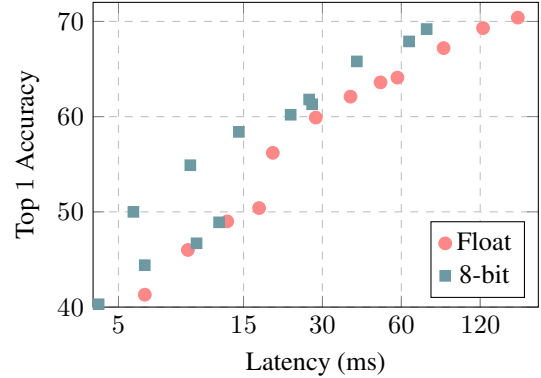
\begin{figure}[h]
\centering
\begin{tikzpicture}
\begin{axis}[
    width=0.9\linewidth,
    height=0.25\textheight,
    compat=1.3,
    xlabel={Latency (ms)},
    ylabel={Top $1$ Accuracy},
    xmin=4, xmax=200,
    ymin=40, ymax=72,
    xtick={5, 15, 30, 60, 120},
    ytick={40, 50, 60, 70},
    legend pos=south east,
    ymajorgrids=true,
    xmajorgrids=true,
    xmode=log,
    log ticks with fixed point,
    grid style=dashed,
]
 
\addplot[
    only marks,
    color=plotred,
    mark=*,
    mark options={scale=1.2},
    ]
    coordinates {
(167,70.4)(123,69.3)(87,67.2)(58,64.1)(50,63.6)(38.4,62.1)(28.3,59.9)(19.4,56.2)(17.2,50.4)(13,49)(9.2,46)(6.3,41.3)
    };
\addplot[
    only marks,
    color=plotgreen,
    mark=square*,
    ]
    coordinates {
(75,69.2)(64.2,67.9)(40.6,65.8)(26.7,61.8)(27.4,61.3)(22.7,60.2)(14.4,58.4)(9.4,54.9)(12.1,48.9)(9.93,46.7)(6.3,44.4)(4.2,40.3)(5.7,50)
};
    \legend{ Float, 8-bit}
 
\end{axis}
\end{tikzpicture}
\centering
\caption{ImageNet classifier on Qualcomm Snapdragon 835 big cores: Latency-vs-accuracy tradeoff of floating-point and integer-only MobileNets.}
\label{fig:imagenet_pixel_2_s835_big}
\end{figure}

\begin{figure}[h]
\centering
\begin{tikzpicture}
\begin{axis}[
    compat=1.3,
    width=0.9\linewidth,
    height=0.25\textheight,
    xlabel={Latency (ms)},
    ylabel={Top $1$ Accuracy},
    xmin=4, xmax=200,
    ymin=40, ymax=72,
    xtick={5, 15, 30, 60, 120},
    ytick={40, 50, 60, 70},
    legend pos=south east,
    ymajorgrids=true,
    xmajorgrids=true,
    xmode=log,
    log ticks with fixed point,
    grid style=dashed,
]
 
\addplot[
    only marks,
    color=plotred,
    mark=*,
    ]
    coordinates {
(120,70.4)(90.5,69.3)(64.3,67.2)(45.1,64.1)(36,63.6)(26.9,62.1)(18.8,59.9)(12.9,56.2)(12.4,50.4)(9.17,49)(6.7,46)(5.8,41.3)
    };
\addplot[
    only marks,
    color=plotgreen,
    mark=square*,
    ]
    coordinates {
(84.7,69.2)(71.9,67.9)(44.8,65.8)(29.5,61.8)(30,61.3)(24.2,60.2)(15.3,58.4)(10.2,54.9)(11.5,48.9)(9.3,46.7)(4.9,44.4)(4.2,40.3)(6.42,50)
};
    \legend{Float, 8-bit}
 
\end{axis}
\end{tikzpicture}

\centering
\caption{ImageNet classifier on Qualcomm Snapdragon 821: Latency-vs-accuracy tradeoff of floating-point and integer-only MobileNets.}
\label{fig:imagenet_pixel_XL_s821}
\end{figure}
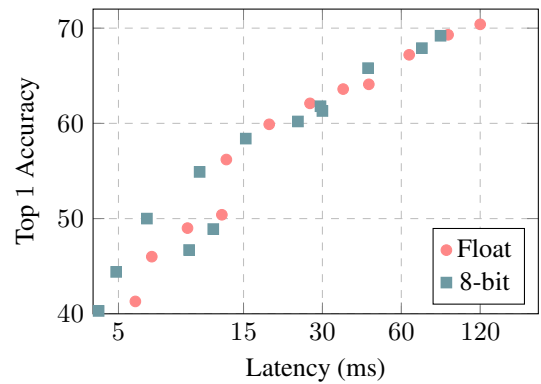

We benchmarked the MobileNet architecture with varying depth-multipliers (DM) and resolutions on ImageNet on three types of Qualcomm cores, which represent three different micro-architectures: 1) Snapdragon 835 LITTLE core,  (figure.~\ref{fig:titlefig}c), a power-efficient processor found in Google Pixel 2; 2) Snapdragon 835 big core (figure.~\ref{fig:imagenet_pixel_2_s835_big}), a high-performance core employed by Google Pixel 2; and 3) Snapdragon 821 big core (figure.~\ref{fig:imagenet_pixel_XL_s821}), a high-performance core used in Google Pixel 1. 


 
 

Integer-only quantized MobileNets achieve higher accuracies than floating-point MobileNets given the same runtime budget. The accuracy gap is quite substantial ($\sim10\%$) for Snapdragon 835 LITTLE cores at the 33ms latency needed for real-time (30 fps) operation. While most of the quantization literature focuses on minimizing accuracy loss for a given architecture, we advocate for a more comprehensive latency-vs-accuracy tradeoff as a better measure. Note that this tradeoff depends critically on the relative speed of floating-point vs integer-only arithmetic in hardware. Floating-point computation is better optimized in the Snapdragon 821, for example, resulting in a less noticeable reduction in latency for quantized models.

\subsubsection{COCO}
We evaluated quantization in the context of mobile real time object detection, comparing the performance of quantized 8-bit and float models of MobileNet SSD \cite{MobilenetV1, liu2015ssd} on the COCO dataset \cite{COCO}. We replaced all the regular convolutions in the SSD prediction layers with separable convolutions (depthwise followed by $1\times1$ projection). This modification is consistent with the overall design of MobileNets and makes them more computationally efficient. We utilized the Open Source TensorFlow Object Detection API \cite{detection-api} to train and evaluate our models. The training protocol is described in Appendix~\ref{sec:coco-protocol}. We also delayed quantization for $500$ thousand steps (see section \ref{sec:quant-range}), finding that it significantly decreases the time to convergence.

Table \ref{tab:coco_detection} shows the latency-vs-accuracy tradeoff between floating-point and integer-quantized models. Latency was measured on a single thread using Snapdragon 835 cores (big and LITTLE). Quantized training and inference results in up to a $50\%$ reduction in running time, with a minimal loss in accuracy ($-1.8\%$ relative).

\begin{table}
\centering
\begin{tabular}{ccccc}
\toprule
    DM    & Type    & mAP   &  LITTLE (ms) & big (ms)     \\
\midrule
    100\% & floats  & 22.1  & 778 & 370 \\
          & 8 bits  & 21.7  & 687 & 272 \\
\midrule
    50\%  & floats  & 16.7  & 270   & 121 \\
          & 8 bits  & 16.6  & 146   & 61 \\
\bottomrule
\end{tabular}
\vspace*{7pt plus 3pt minus 2pt}
\caption{Object detection speed and accuracy on COCO dataset of floating point and integer-only quantized models. Latency (ms) is measured on Qualcomm Snapdragon 835 big and LITTLE cores.}
\label{tab:coco_detection}
\end{table}

\vspace{-.1in}
\subsubsection{Face detection}
To better examine quantized MobileNet SSD on a smaller scale, we benchmarked face detection on the face attribute classification dataset (a Flickr-based dataset used in~\cite{MobilenetV1}). We contacted the authors of~\cite{MobilenetV1} to evaluate our quantized MobileNets on detection and face attributes following the same protocols (detailed in Appendix~\ref{sec:face-protocol}).

As indicated by tables~\ref{tab:face_detection_precision} and~\ref{tab:face_detection_latency}, quantization provides close to a $2\times$ latency reduction with a Qualcomm Snapdragon 835 big or LITTLE core at the cost of a $\sim2\%$ drop in the average precision. Notably, quantization allows the $25\%$ face detector to run in real-time ($1K/28\approx36$ fps) on a single big core, whereas the floating-point model remains slower than real-time ($1K/44\approx23$ fps). 

We additionally examine the effect of multi-threading on the latency of quantized models. Table~\ref{tab:face_detection_latency} shows a $1.5$ to $2.2\times$) speedup when using $4$ cores. The speedup ratios are comparable between the two cores, and are higher for larger models where the overhead of multi-threading occupies a smaller fraction of the total computation.

\begin{table}
\centering
\begin{tabular}{cccc}
\toprule
    DM    & type    & Precision            & Recall \\
\midrule
    100\% & floats  & 68\%  & 76\%   \\
          & 8 bits  & 66\%  & 75\%   \\
\midrule
    50\%  & floats  & 65\%  & 70\%   \\
          & 8 bits  & 62\%  & 70\%   \\
\midrule
    25\%  & floats  & 56\%  & 64\%   \\
          & 8 bits  & 54\%  & 63\%   \\
\bottomrule
\end{tabular}
\vspace*{8pt plus 3pt minus 2pt}
\caption{Face detection accuracy of floating point and integer-only quantized models. The reported precision / recall is averaged over different precision / recall values where an IOU of $x$ between the groundtruth and predicted windows is considered a correct detection, for $x$ in $\{0.5, 0.55, \ldots, 0.95\}$. }
\label{tab:face_detection_precision}
\end{table}

\begin{table}
\centering
\begin{tabular}{cccccccc}
\toprule
    DM    & type    & \multicolumn{3}{c}{LITTLE Cores} & \multicolumn{3}{c}{big Cores} \\
          &         & 1   & 2   & 4              & 1   & 2   & 4 \\

\midrule
    100\% & floats  & 711 & --  & --             & 337 & --  & --   \\
          & 8 bits  & 372 & 238 & 167            & 154 & 100 & 69   \\
\midrule
    50\%  & floats  & 233 & --  & --             & 106 & --  & --   \\
          & 8 bits  & 134 & 96  & 74             & 56  & 40  & 30   \\
\midrule
    25\%  & floats  & 100 & --  & --             & 44  & --  & --   \\
          & 8 bits  & 67  & 52  & 43             & 28  & 22  & 18   \\
\bottomrule
\end{tabular}
\vspace*{8pt plus 3pt minus 2pt}
\caption{Face detection: latency of floating point and quantized models on Qualcomm Snapdragon 835 cores.}
\label{tab:face_detection_latency}
\end{table}

\subsubsection{Face attributes}
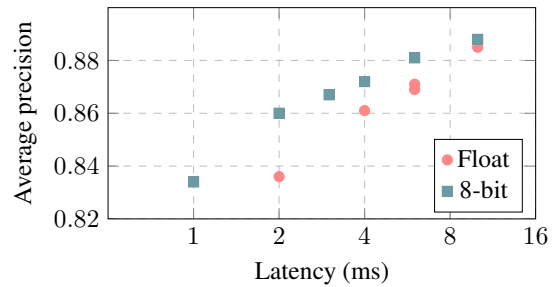
\begin{figure}[h]
\resizebox{0.9\columnwidth}{!}{
\begin{tikzpicture}
\begin{axis}[
    width=0.9\linewidth,
    height=0.2\textheight,
compat=1.3,
    xlabel={Latency (ms)},
    ylabel={Average precision},
    xmin=0.5, xmax=16,
    ymin=0.82, ymax=0.9,
    xtick={1, 2, 4, 8, 16},
    ytick={0.82, 0.84, 0.86, 0.88},
    legend pos=south east,
    ymajorgrids=true,
    xmajorgrids=true,
    xmode=log,
    log ticks with fixed point,
    grid style=dashed,
]
 
\addplot[
    only marks,
    color=plotred,
    mark=*,
    ]
    coordinates {
(6,0.871)(4,0.861)(2,0.836)(17,0.889)(10,0.885)(6,0.869)
    };
\addplot[
    only marks,
    color=plotgreen,
    mark=square*,
    ]
    coordinates {
(4,0.872)(2,0.860)(1,0.834)(10,0.888)(6,0.881)(3,0.867)
};
    \legend{Float, 8-bit}
 
\end{axis}
\end{tikzpicture}
}
\centering
\caption{Face attribute classifier on Qualcomm Snapdragon 821: Latency-vs-accuracy tradeoff of floating-point and integer-only MobileNets.}
\label{fig:fa_pixel_XL_s821}
\end{figure}

Figure \ref{fig:fa_pixel_XL_s821} shows the latency-vs-accuracy tradeoff of face attribute classification on the Qualcomm Snapdragon 821. Since quantized training results in little accuracy degradation, we see an improved tradeoff even though the Qualcomm Snapdragon 821 is highly optimized for floating point arithmetic (see Figure~\ref{fig:imagenet_pixel_XL_s821} for comparison).

\textbf{Ablation study} To understand performance sensitivity to the quantization scheme, we further evaluate quantized training with varying weight and activation quantization bit depths. The degradation in average precision for binary attributes and age precision relative to the floating-point baseline are shown in
Tables \ref{tab:face_attr_avg_map} and \ref{tab:face_attr_age_precision_5}, respectively. The tables suggest that 1) weights are more sensitive to reduced quantization bit depth than activations, 2) 8 and 7-bit quantized models perform similarly to floating point models, and 3) when the total bit-depths are equal, it is better to keep weight and activation bit depths the same.

\begin{table}
\centering
\resizebox{\columnwidth}{!}{
\begin{tabular}{cccccc}
\toprule
    \diagbox{wt.}{act.} & 8 & 7 & 6 & 5 & 4 \\
\midrule
    8 &  -0.9\% &  -0.3\% &  -0.4\% &  -1.3\% &  -3.5\% \\
    7 &  -1.3\% &  -0.5\% &  -1.2\% &  -1.0\% &  -2.6\% \\
    6 &  -1.1\% &  -1.2\% &  -1.6\% &  -1.6\% &  -3.1\% \\
    5 &  -3.1\% &  -3.7\% &  -3.4\% &  -3.4\% &  -4.8\% \\
    4 & -11.4\% & -13.6\% & -10.8\% & -13.1\% & -14.0\% \\
\bottomrule
\end{tabular}
}
\vspace*{8pt plus 3pt minus 2pt}
\caption{Face attributes: relative average category precision of integer-quantized MobileNets (varying weight and activation bit depths) compared with floating point.}
\label{tab:face_attr_avg_map}
\end{table}

\begin{table}
\centering
\resizebox{\columnwidth}{!}{
\begin{tabular}{cccccc}
\toprule
    \diagbox{wt.}{act.} & 8 & 7 & 6 & 5 & 4 \\
\midrule
    8 &  -1.3\% &  -1.6\% &  -3.2\% &  -6.0\% &  -9.8\% \\
    7 &  -1.8\% &  -1.2\% &  -4.6\% &  -7.0\% &  -9.9\% \\
    6 &  -2.1\% &  -4.9\% &  -2.6\% &  -7.3\% &  -9.6\% \\
    5 &  -3.1\% &  -6.1\% &  -7.8\% &  -4.4\% & -10.0\% \\
    4 & -10.6\% & -20.8\% & -17.9\% & -19.0\% & -19.5\% \\
\bottomrule
\end{tabular}
}
\vspace*{8pt plus 3pt minus 2pt}
\caption{Face attributes: Age precision at difference of 5 years for quantized model (varying weight and activation bit depths) compared with floating point.}
\label{tab:face_attr_age_precision_5}
\end{table}

\section{Discussion}
We propose a quantization scheme that relies only on integer arithmetic to approximate the floating-point computations in a neural network. Training that simulates the effect of quantization helps to restore model accuracy to near-identical levels as the original. In addition to the $4\times$ reduction of model size, inference efficiency is improved via ARM NEON-based implementations. The improvement advances the state-of-the-art tradeoff between latency on common ARM CPUs and the accuracy of popular computer vision models. The synergy between our quantization scheme and efficient architecture design suggests that integer-arithmetic-only inference could be a key enabler that propels visual recognition technologies into the real-time and low-end phone market.

{\small
\bibliographystyle{ieee}
\bibliography{bibliography}
}

\clearpage
\appendix
\renewcommand{\theequation}{A\thechapter.\arabic{equation}}
\numberwithin{equation}{section}
\section{Appendix: Layer-specific details}

\subsection{Mathematical functions}\label{sec:math-functions}

Math functions such as hyperbolic tangent, the logistic function, and softmax often appear in neural networks.
No lookup tables are needed since these functions are implemented in pure fixed-point arithmetic similarly to how they would be implemented in floating-point arithmetic\footnote{Pure-arithmetic, SIMD-ready, branch-free, fixed-point implementations of at least tanh and the logistic functions are given in gemmlowp \cite{gemmlowp}'s \href{https://github.com/google/gemmlowp/tree/fcf32e7a0a4d2af46e63eccf0c8fa4d83d0311c5/fixedpoint}{fixedpoint directory}, with specializations for NEON and SSE instruction sets. One can see in TensorFlow Lite \cite{tflite} \href{https://github.com/tensorflow/tensorflow/blob/4952f981be07b8bf508f8226f83c10cdafa3f0c4/tensorflow/contrib/lite/kernels/internal/optimized/optimized_ops.h\#L2705-L2844}{how these are called}.}.

\subsection{Addition}

Some neural networks use a plain Addition layer type, that simply adds two activation arrays together. Such Addition layers are more expensive in quantized inference compared to floating-point because \emph{rescaling} is needed: one input needs to be rescaled onto the other's scale using a fixed-point multiplication by the multiplier $M=S_1/S_2$ similar to what we have seen earlier (end of section \ref{sec:integer-only-inference}), before the actual addition can be performed as a simple integer addition; finally, the result must be rescaled again to fit the output array's scale\footnote{See the TensorFlow Lite \cite{tflite} \href{https://github.com/tensorflow/tensorflow/blob/4952f981be07b8bf508f8226f83c10cdafa3f0c4/tensorflow/contrib/lite/kernels/internal/optimized/optimized_ops.h\#L1402-L1507}{implementation}.}.

\subsection{Concatenation}

Fully general support for concatenation layers poses the same rescaling problem as Addition layers. Because such rescaling of uint8 values would be a lossy operation, and as it seems that concatenation ought to be a lossless operation, we prefer to handle this problem differently: instead of implementing lossy rescaling, we introduce a requirement that all the input activations and the output activations in a Concatenation layer have the same quantization parameters. This removes the need for rescaling and concatenations are thus lossless and free of any arithmetic\footnote{This is implemented in \href{https://github.com/tensorflow/tensorflow/blob/faf7f05f5ed3d92405656a318fb2d571a7d31532/tensorflow/contrib/lite/toco/graph_transformations/hardcode_min_max.cc\#L66-L126}{this part} of the TensorFlow Lite \cite{tflite} Converter}.

\section{Appendix: ARM NEON details}\label{sec:arm-neon}

This section assumes familiarity with assembly programming on the ARM NEON instruction set. The instruction mnemonics below refer to the 64-bit ARM instruction set, but the discussion applies equally to 32-bit ARM instructions.

The fixed-point multiplications referenced throughout this article map exactly to the SQRDMULH instruction. It is very important to use the correctly-rounding instruction SQRDMULH and not SQDMULH\footnote{The fixed-point math function \href{https://github.com/google/gemmlowp/tree/fcf32e7a0a4d2af46e63eccf0c8fa4d83d0311c5/fixedpoint}{implementations} in gemmlowp \cite{gemmlowp} use such fixed-point multiplications, and ordinary (non-saturating) integer additions. We have no use for general saturated arithmetic.}.

The rounding-to-nearest right-shifts referenced in section \ref{sec:integer-only-inference} do not map exactly to any ARM NEON instruction. The problem is that the ``rounding right shift" instruction, RSHL with variable negative offset, breaks ties by rounding \emph{upward}, instead of rounding them \emph{away from zero}. For example, if we use RSHL to implement the division $-12/2^3$, the result will be $-1$ whereas it should be $-2$ with ``round to nearest". This is problematic as it results in an overall upward bias, which has been observed to cause significant loss of end-to-end accuracy in neural network inference. A correct round-to-nearest right-shift can still be implemented using RSHL but with suitable fix-up arithmetic around it\footnote{It is implemented \href{https://github.com/google/gemmlowp/blob/fcf32e7a0a4d2af46e63eccf0c8fa4d83d0311c5/fixedpoint/fixedpoint_neon.h\#L146-L152}{here} in gemmlowp \cite{gemmlowp}.}.

For efficient NEON implementation of the matrix multiplication's core accumulation, we use the following trick. In the multiply-add operation in (\ref{eq:accum-types}), we first change the operands' type from uint8 to int8 (which can be done by subtracting 128 from the quantized values and zero-points). Thus the core multiply-add becomes
\begin{equation}
\label{eq:accum-types-signed}
\texttt{int32 += int8 * int8}.
\end{equation}
As mentioned in section \ref{sec:training}, with a minor tweak of the quantized training process, we can ensure that the weights, once quantized as int8 values, never take the value $-128$. Hence, the product in (\ref{eq:accum-types-signed}) is never $-128*-128$, and is therefore always less than $2^{14}$ in absolute value. Hence, (\ref{eq:accum-types-signed}) can accumulate \emph{two} products on a local int16 accumulator before that needs to be accumulated into the true int32 accumulator. This allows the use of an 8-way SIMD multiplication (SMULL on int8 operands), followed by an 8-way SIMD multiply-add (SMLAL on int8 operands), followed by a pairwise-add-and-accumulate into the int32 accumulators (SADALP)\footnote{This technique is implemented in the \href{https://github.com/google/gemmlowp/blob/fcf32e7a0a4d2af46e63eccf0c8fa4d83d0311c5/internal/kernel_neon.h\#L929-L1262}{optimized NEON kernel} in gemmlowp \cite{gemmlowp}, which is in particular what TensorFlow Lite uses (see the choice of \texttt{L8R8WithLhsNonzeroBitDepthParams} at \href{https://github.com/tensorflow/tensorflow/blob/4952f981be07b8bf508f8226f83c10cdafa3f0c4/tensorflow/contrib/lite/kernels/internal/optimized/optimized_ops.h\#L903}{this line}).}.

\section{Appendix: Graph diagrams}\label{sec:tf_graphs}

\begin{figure}[h]
\centering
\begin{tikzpicture}[node distance=2cm,scale=0.7,transform shape]
\node (conv) [conv] {conv};
\node (weights) [params, below right of=conv] {weights};
\node (input) [nobox, below left of=weights] {input};
\node (add) [add, above of=conv] {+};
\node (biases) [params, below left of=add, xshift=-0.8cm] {biases};
\node (ReLU6) [act, above of=add] {ReLU6};
\node (output) [nobox, above of=ReLU6] {output};
\draw [arrow] (input) -- (conv);
\draw [arrow] (weights) -- (conv);
\draw [arrow] (conv) -- (add);
\draw [arrow] (biases) -- (add);
\draw [arrow] (add) -- (ReLU6);
\draw [arrow] (ReLU6) -- (output);
\end{tikzpicture}
\caption{Simple graph: original}
\label{fig:simple_original}
\end{figure}

\begin{figure}[h]
\centering
\begin{tikzpicture}[node distance=2cm,scale=0.7,transform shape]
\node (conv) [conv] {conv};
\node (wt_quant) [quant, below right of=conv] {wt quant};
\node (weights) [params, below of=wt_quant] {weights};
\node (input) [nobox, below left of=weights] {input};
\node (add) [add, above of=conv] {+};
\node (biases) [params, below left of=add, xshift=-0.8cm] {biases};
\node (ReLU6) [act, above of=add] {ReLU6};
\node (act_quant) [quant, above of=ReLU6] {act quant};
\node (output) [nobox, above of=act_quant] {output};
\draw [arrow] (input) -- (conv);
\draw [arrow] (weights) -- (wt_quant);
\draw [arrow] (wt_quant) -- (conv);
\draw [arrow] (conv) -- (add);
\draw [arrow] (biases) -- (add);
\draw [arrow] (add) -- (ReLU6);
\draw [arrow] (ReLU6) -- (act_quant);
\draw [arrow] (act_quant) -- (output);
\end{tikzpicture}
\caption{Simple graph: quantized}
\label{fig:simple_quantized}
\end{figure}
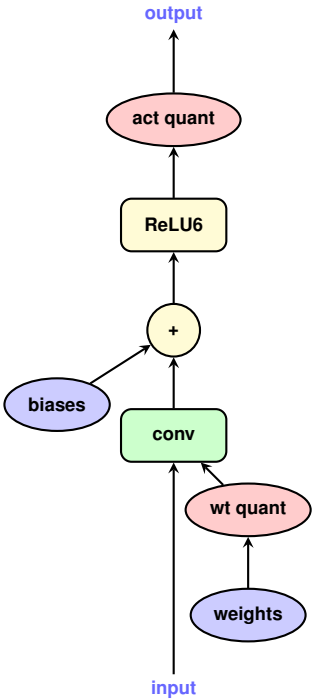

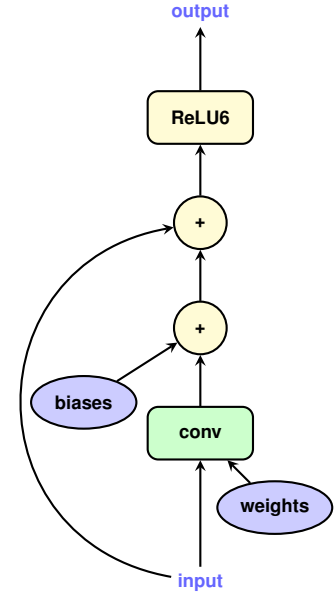
\begin{figure}[h]
\centering
\begin{tikzpicture}[node distance=2cm,scale=0.7,transform shape]
\node (conv) [conv] {conv};
\node (weights) [params, below right of=conv] {weights};
\node (input) [nobox, below left of=weights] {input};
\node (add) [add, above of=conv] {+};
\node (biases) [params, below left of=add, xshift=-0.8cm] {biases};
\node (add2) [add, above of=add] {+};
\node (ReLU6) [act, above of=add2] {ReLU6};
\node (output) [nobox, above of=ReLU6] {output};
\draw [arrow] (input) -- (conv);
\draw [arrow] (weights) -- (conv);
\draw [arrow] (conv) -- (add);
\draw [arrow] (biases) -- (add);
\draw [arrow] (add) -- (add2);
\draw [arrow] (input) to [out=170,in=190,looseness=1.5] (add2);
\draw [arrow] (add2) -- (ReLU6);
\draw [arrow] (ReLU6) -- (output);
\end{tikzpicture}
\caption{Layer with a bypass connection: original}
\label{fig:bypass_original}
\end{figure}

\begin{figure}[h]
\centering
\begin{tikzpicture}[node distance=2cm,scale=0.7,transform shape]
\node (conv) [conv] {conv};
\node (wt_quant) [quant, below right of=conv] {wt quant};
\node (weights) [params, below of=wt_quant] {weights};
\node (input) [nobox, below left of=weights] {input};
\node (add) [add, above of=conv] {+};
\node (biases) [params, below left of=add, xshift=-0.8cm] {biases};
\node (conv_quant) [quant, above of=add] {conv quant};
\node (add2) [add, above of=conv_quant] {+};
\node (ReLU6) [act, above of=add2] {ReLU6};
\node (act_quant) [quant, above of=ReLU6] {act quant};
\node (output) [nobox, above of=act_quant] {output};
\draw [arrow] (input) -- (conv);
\draw [arrow] (weights) -- (wt_quant);
\draw [arrow] (wt_quant) -- (conv);
\draw [arrow] (conv) -- (add);
\draw [arrow] (biases) -- (add);
\draw [arrow] (add) -- (conv_quant);
\draw [arrow] (conv_quant) -- (add2);
\draw [arrow] (input) to [out=160,in=200] (add2);
\draw [arrow] (add2) -- (ReLU6);
\draw [arrow] (ReLU6) -- (act_quant);
\draw [arrow] (act_quant) -- (output);
\end{tikzpicture}
\caption{Layer with a bypass connection: quantized}
\label{fig:bypass_quantized}
\end{figure}
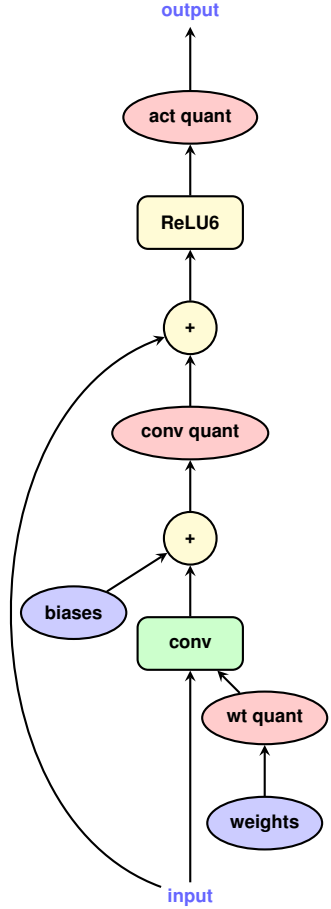

\begin{figure}[h]
\centering
\begin{tikzpicture}[node distance=2cm,scale=0.7,transform shape]
\node (conv) [conv] {conv};
\node (weights) [params, below right of=conv] {weights};
\node (input) [nobox, below left of=weights] {input};
\node (moments) [act, above right of=conv] {moments};
\node (ma) [ma, above of=moments] {MA $\mu$, $\sigma$};
\node (bn_add) [act, above left of=ma] {$\gamma (x - \mu) / \sigma  + \beta$};
\node (gamma) [small_params, below left of=bn_add] {$\gamma$};
\node (beta) [small_params, below left of=bn_add, xshift=-1.5cm] {$\beta$};
\node (ReLU6) [act, above of=bn_add] {ReLU6};
\node (output) [nobox, above of=ReLU6] {output};
\draw [arrow] (input) -- (conv);
\draw [arrow] (weights) -- (conv);
\draw [arrow] (conv) -- (bn_add);
\draw [arrow] (conv) -- (moments);
\draw [arrow] (moments) -- (ma);
\draw [arrow] (ma) -- (bn_add);
\draw [arrow] (gamma) -- (bn_add);
\draw [arrow] (beta) -- (bn_add);
\draw [arrow] (bn_add) -- (ReLU6);
\draw [arrow] (ReLU6) -- (output);
\end{tikzpicture}
\caption{Convolutional layer with batch normalization: training graph}
\label{fig:bn_train}
\end{figure}

\begin{figure}[h]
\centering
\begin{tikzpicture}[node distance=2cm,scale=0.7,transform shape]
\node (conv) [conv] {conv};
\node (weights) [params, below right of=conv] {$w \gamma / \sigma$};
\node (input) [nobox, below left of=weights] {input};
\node (add) [add, above of=conv] {+};
\node (biases) [params, below left of=add, xshift=-1cm] {$\beta - \gamma \mu / \sigma$};
\node (ReLU6) [act, above of=add] {ReLU6};
\node (output) [nobox, above of=ReLU6] {output};
\draw [arrow] (input) -- (conv);
\draw [arrow] (weights) -- (conv);
\draw [arrow] (conv) -- (add);
\draw [arrow] (biases) -- (add);
\draw [arrow] (add) -- (ReLU6);
\draw [arrow] (ReLU6) -- (output);
\end{tikzpicture}
\caption{Convolutional layer with batch normalization: inference graph}
\label{fig:bn_inference}
\end{figure}
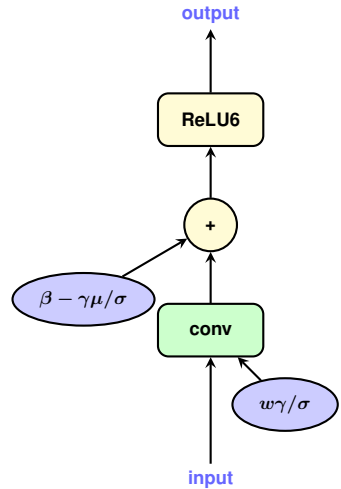

\begin{figure}[h]
\centering
\begin{tikzpicture}[node distance=2cm,scale=0.7,transform shape]
\node (conv) [conv] {conv};
\node (weights) [params, below right of=conv] {weights};
\node (input) [nobox, below left of=weights] {input};
\node (moments) [act, above of=conv] {moments};
\node (ma) [ma, above of=moments] {MA $\mu$, $\sigma$};
\node (fold_weights) [act, above left of=ma] {$w \gamma / \sigma$};
\node (gamma) [small_params, below left of=fold_weights] {$\gamma$};
\node (fold_conv) [conv, above of=fold_weights] {conv fold};
\node (fold_biases) [act, above right of=ma] {$\beta - \gamma \mu / \sigma$};
\node (beta) [small_params, below right of=fold_biases] {$\beta$};
\node (add) [add, above of=fold_conv] {+};
\node (ReLU6) [act, above of=add] {ReLU6};
\node (output) [nobox, above of=ReLU6] {output};
\draw [arrow] (input) -- (conv);
\draw [arrow] (weights) -- (conv);
\draw [arrow] (conv) -- (moments);
\draw [arrow] (moments) -- (ma);
\draw [arrow] (ma) -- (fold_weights);
\draw [arrow] (gamma) -- (fold_weights);
\draw [arrow] (ma) -- (fold_biases);
\draw [arrow] (beta) -- (fold_biases);
\draw [arrow] (fold_weights) -- (fold_conv);
\draw [arrow] (input) to [out=135,in=225,looseness=1.1] (fold_conv);
\draw [arrow] (fold_conv) -- (add);
\draw [arrow] (fold_biases) -- (add);
\draw [arrow] (add) -- (ReLU6);
\draw [arrow] (ReLU6) -- (output);
\end{tikzpicture}
\caption{Convolutional layer with batch normalization: training graph, folded}
\label{fig:bn_fold_train}
\end{figure}
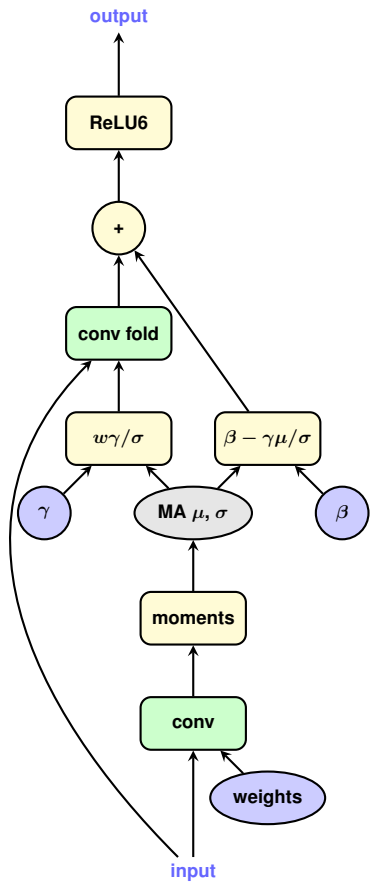

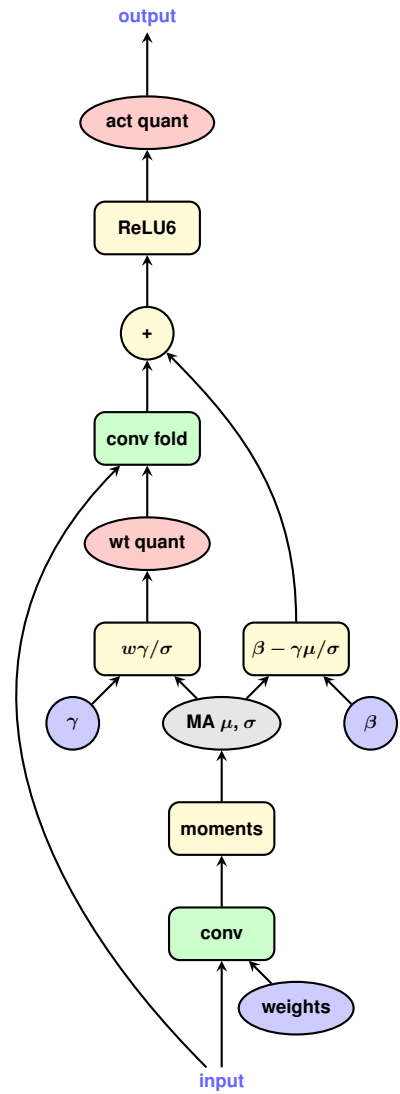
\begin{figure}[h]
\centering
\begin{tikzpicture}[node distance=2cm,scale=0.7,transform shape]
\node (conv) [conv] {conv};
\node (weights) [params, below right of=conv] {weights};
\node (input) [nobox, below left of=weights] {input};
\node (moments) [act, above of=conv] {moments};
\node (ma) [ma, above of=moments] {MA $\mu$, $\sigma$};
\node (fold_weights) [act, above left of=ma] {$w \gamma / \sigma$};
\node (gamma) [small_params, below left of=fold_weights] {$\gamma$};
\node (wt_quant) [quant, above of=fold_weights] {wt quant};
\node (fold_conv) [conv, above of=wt_quant] {conv fold};
\node (fold_biases) [act, above right of=ma] {$\beta - \gamma \mu / \sigma$};
\node (beta) [small_params, below right of=fold_biases] {$\beta$};
\node (add) [add, above of=fold_conv] {+};
\node (ReLU6) [act, above of=add] {ReLU6};
\node (act_quant) [quant, above of=ReLU6] {act quant};
\node (output) [nobox, above of=act_quant] {output};
\draw [arrow] (input) -- (conv);
\draw [arrow] (weights) -- (conv);
\draw [arrow] (conv) -- (moments);
\draw [arrow] (moments) -- (ma);
\draw [arrow] (ma) -- (fold_weights);
\draw [arrow] (gamma) -- (fold_weights);
\draw [arrow] (ma) -- (fold_biases);
\draw [arrow] (beta) -- (fold_biases);
\draw [arrow] (fold_weights) -- (wt_quant);
\draw [arrow] (wt_quant) -- (fold_conv);
\draw [arrow] (input) to [out=135,in=225,looseness=1.1] (fold_conv);
\draw [arrow] (fold_conv) -- (add);
\draw [arrow] (fold_biases) to [out=90,in=315] (add);
\draw [arrow] (add) -- (ReLU6);
\draw [arrow] (ReLU6) -- (act_quant);
\draw [arrow] (act_quant) -- (output);
\end{tikzpicture}
\caption{Convolutional layer with batch normalization: training graph, folded and quantized}
\label{fig:bn_fold_quant_train}
\end{figure}

\section{Experimental protocols}\label{sec:protocols}
\subsection{ResNet protocol}\label{sec:resnet-protocol}
\textbf{Preprocessing}. All images from ImageNet \cite{deng2009imagenet} are resized preserving aspect ratio so that the smallest side of the image is $256$. Then the center $224\times224$ patch is cropped and the means are subtracted for each of the RGB channels.

\textbf{Optimization}. We use the momentum optimizer from TensorFlow \cite{abadi2015tensorflow} with momentum $0.9$ and a batch size of $32$. The learning rate starts from $10^{-5}$ and decays in a staircase fashion by $0.1$ for every $30$ epochs. Activation quantization is delayed for $500,000$ steps for reasons discussed in section \ref{sec:training}. Training uses $50$ workers asynchronously, and stops after validation accuracy plateaus, normally after $100$ epochs. 

\subsection{Inception protocol}\label{sec:inception-protocol}
All results in table \ref{tab:inception_v3_imagenet} were obtained after training for approximately $10$ million steps, with batches of $32$ samples, using $50$ distributed workers, asynchronously.  Training data were ImageNet 2012 $299 \times 299$ images with labels. Image augmentation consisted of: random crops, random horizontal flips, and random color distortion. The optimizer used was RMSProp with learning rate starting at $0.045$ and decaying exponentially and stepwise with factor $0.94$ after every $2$ epochs.  Other RMSProp parameters were: $0.9$ momentum, $0.9$ decay, $1.0$ epsilon term.  Trained parameters were EMA averaged with decay $0.9999$.

\subsection{COCO detection protocol}\label{sec:coco-protocol}

\textbf{Preprocessing}. During training, all images are randomly cropped and resized to $320\times320$. During evaluation, all images are directly resized to $320\times320$. All input values are normalized to $[-1, 1]$.

\textbf{Optimization}.  We used the RMSprop optimizer from TensorFlow \cite{abadi2015tensorflow} with a batch size of $32$. The learning rate starts from $4\times10^{-3}$ and decays in a staircase fashion by a factor of $0.1$ for every $100$ epochs. Activation quantization is delayed for $500,000$ steps for reasons discussed in section \ref{sec:training}. Training uses $20$ workers asynchronously, and stops after validation accuracy plateaus, normally after approximately $6$ million steps.

\textbf{Metrics}. Evaluation results are reported with the COCO primary challenge metric: AP at IoU=.50:.05:.95. We follow the same train/eval split in \cite{cocodetection2016}.

\subsection{Face detection and face attribute classification protocol}\label{sec:face-protocol}

\textbf{Preprocessing}. Random 1:1 crops are taken from images in the Flickr-based dataset used in~\cite{MobilenetV1} and resized to $320\times320$ pixels for face detection and $128\times128$ pixels for face attribute classification. The resulting crops are flipped horizontally with a $50\%$ probability. The values for each of the RGB channels are renormalized to be in the range $[-1, 1]$.

\textbf{Face Detection Optimization}. We used the RMSprop optimizer from TensorFlow \cite{abadi2015tensorflow} with a batch size of $32$. The learning rate starts from $4\times10^{-3}$ and decays in a staircase fashion by a factor of $0.1$ for every $100$ epochs. Activation quantization is delayed for $500,000$ steps for reasons discussed in section \ref{sec:training}. Training uses $20$ workers asynchronously, and stops after validation accuracy plateaus, normally after approximately $3$ million steps.

\textbf{Face Attribute Classification Optimization}. We followed the optimization protocol in~\cite{MobilenetV1}. We used the Adagrad optimizer from Tensorflow\cite{abadi2015tensorflow} with a batch size of $32$ and a constant learning rate of $0.1$. Training uses $12$ workers asynchronously, and stops at $20$ million steps.

\textbf{Latency Measurements}. We created a binary that runs the face detection and face attributes classification models repeatedly on random inputs for $100$ seconds. We pushed this binary to Pixel and Pixel 2 phones using the \texttt{adb push} command, and executed it on 1, 2, and 4 LITTLE cores, and 1, 2, and 4 big cores using the \texttt{adb shell} command with the appropriate \texttt{taskset} specified. We reported the average runtime of the face detector model on $320\times320$ inputs, and of the face attributes classifier model on $128\times128$ inputs.

\end{document}